%% file: main.tex
\pdfoutput=1

\documentclass[11pt]{article}

\usepackage[]{acl} 
\usepackage{times}
\usepackage{latexsym}
\usepackage{booktabs}
\usepackage{amsmath}
\usepackage{subcaption} 
\usepackage{multirow}

\usepackage[T1]{fontenc}

\usepackage[utf8]{inputenc}

\usepackage{microtype}
\usepackage{graphicx}

\usepackage{inconsolata}

\title{Aged to Perfection: Machine-Learning Maps of Age in Conversational English}

\author{MingZe Tang\\
  Department of Computing Science \\
  University of Aberdeen \\
  52426282 \\
  \texttt{m.tang.24\@abdn.ac.uk} \\}

\begin{document}
\maketitle
\begin{abstract}
The study uses the British National Corpus 2014, a large sample of contemporary spoken British English, to investigate language patterns across different age groups. Our research attempts to explore how language patterns vary between different age groups, exploring the connection between speaker demographics and linguistic factors such as utterance duration, lexical diversity, and word choice. By merging computational language analysis and machine learning methodologies, we attempt to uncover distinctive linguistic markers characteristic of multiple generations and create prediction models that can consistently estimate the speaker's age group from various aspects. This work contributes to our knowledge of sociolinguistic diversity throughout the life of modern British speech.
\end{abstract}

\section{Introduction}
Language choices vary systematically across the lifespan, yet quantitative evidence for how these differences pattern in everyday British conversation remains sparse. Leveraging the spoken component of the British National Corpus 2014 with $\approx$1.2k dialogues spanning speakers aged 2–91. This study combines corpus statistics with machine-learning models to trace age-linked shifts in utterance length, lexical diversity, topic salience and semantic similarity. $\chi^2$ testing, embedding-based clustering and a suite of supervised classifiers (Logistic Regression, Random Forest, Gradient Boosting, Linear SVM, MLP) are deployed to isolate diagnostic features and assess the separability of age groups. The analysis offers both a descriptive portrait of life-span linguistic change and an evaluation of how reliably an algorithm can infer a speaker’s age cohort from text and other lexical features.


\input{sections/data_method}
\input{sections/results}

\input{sections/discussion}

\bibliography{acl}
\input{sections/appendix}

\end{document}

%% file: sections/data_method.tex
\section{Description of Data and Methods}

\input{sections/subsections/sub_data}
\input{sections/subsections/sub_method}

%% file: sections/subsections/sub_data.tex
\subsection{Data}
\label{subsec:data}
The British National Corpus \citep{BNC2014} comprises 1,251 spoken-language interviews conducted between 2012 and 2014, featuring speakers aged from 2 to 91 years and rich metadata on demographics (age, gender, education, socioeconomic status) and interactional context. The dialogues are conducted predominantly in a small group of two to four participants with a mean of $\approx$941 tokens per interview, ranging from 61 to 15690 tokens, and an average of $\approx$3 speakers per dialogue as shown in Table \ref{tab:data_stats}. At the turn level, utterances are brief (mean 8.76 tokens; median=5), reflecting a mix of brief acknowledgments and occasional extended narratives.

\begin{table}[h!]
    \centering
    \resizebox{\columnwidth}{!}{%
    \begin{tabular}{lrrrrrrr}
    \toprule
    \textbf{Feature}
      & \textbf{Min}
      & $\mathbf{P}_{\mathbf{25}}$
      & \textbf{Median}
      & $\mathbf{P}_{\mathbf{75}}$
      & \textbf{Max}
      & \textbf{Mean}
      & \textbf{Std} \\
    \midrule
    Speaker Age
      & 2.0
      & 22.0
      & 34.0
      & 55.25
      & 91.0
      & 39.89
      & 20.16 \\
    Utterance Length (Token)
      & 1.0
      & 2.0
      & 5.0
      & 11.0
      & 960.0
      & 8.76
      & 12.36 \\
    Dialogue Length (Utterance)
      & 61.0
      & 399.0
      & 678.0
      & 1225.0
      & 15690.0
      & 941.81
      & 890.98 \\
    Speakers per Dialogue
      & 2.0
      & 2.0
      & 3.0
      & 3.0
      & 12.0
      & 2.87
      & 1.15 \\
    \bottomrule
    \end{tabular}}
    \caption{Descriptive Statistics for Speaker and Dialogue Features, Min, 25th percentile ($P_{25}$), Median, 75th percentile ($P_{75}$), Max, Mean, and Std Dev.}
    \label{tab:data_stats}
\end{table}

\begin{figure}[ht]
    \centering
    \includegraphics[width=\linewidth]{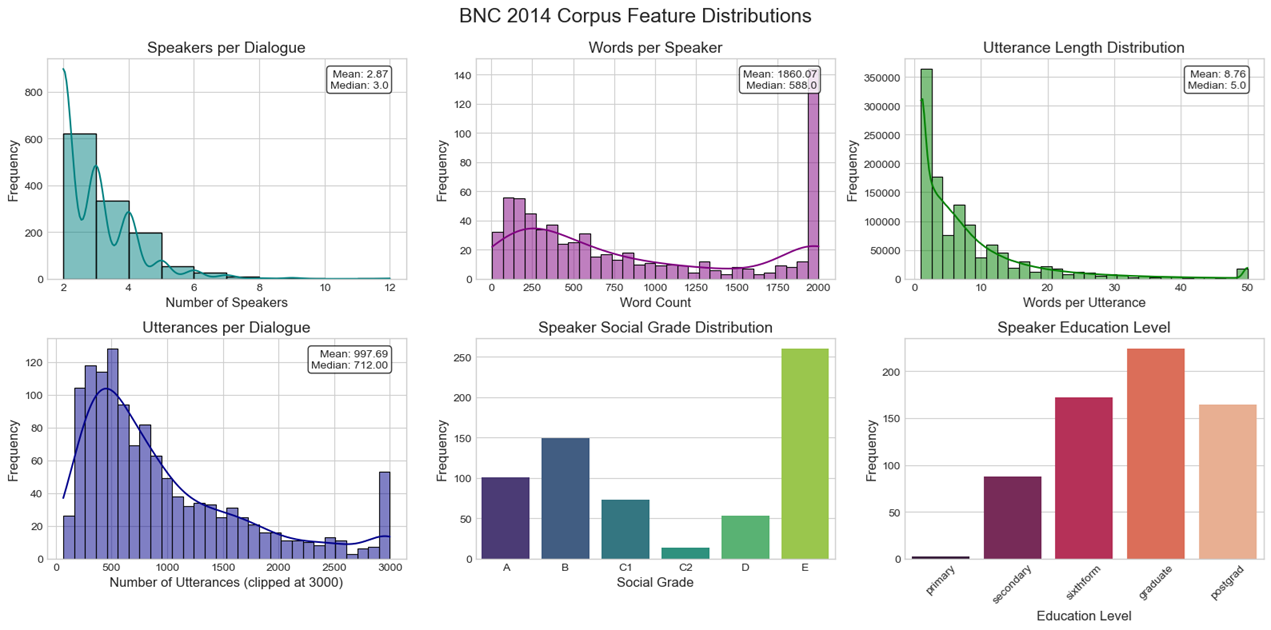}
    \caption{Corpus Feature Distribution (frequency): (a) Speakers per Dialogue, (b) Words per Speaker, (c) Utterance Length, (d) Utterances per Dialogue, (e) Speaker Social Grade, (f) Speaker Education Level}
    \label{fig:corpus_dist}
\end{figure}

The verbosity of the speaker also shows a pronounced skew, as a minority of individuals produce up to $\approx$2000 total tokens per speaker, as shown in Figure \ref{fig:corpus_dist}. Demographically, participants are disproportionately drawn from social grades A-B and from graduate‐educated backgrounds, reflecting an overrepresentation of higher socioeconomic and educational strata.

We divided speakers into five age bins (Kids: 0-12, Teens: 13-19, Young Adults: 20-35, Adults: 36-55, and Seniors: 56+) to simplify comparative examination of language trends across the lifespan. An analysis as shown in Figure \ref{fig:mean_length_age_gender} has shown that the mean utterance length increases significantly with age group $(F(4, 518)=2.97, p=.019)$, rising from roughly $\approx$6 words in the 0–12 year cohort to about $\approx$10 words in speakers over 20. Neither the speaker gender $(t(526)= -1.51, p=.13)$ nor the age-gender interaction $(F(4, 518)=0.91, p=.46)$ reached significance, showing utterance length varies mainly by age rather than by gender.

\begin{figure}[ht]
  \centering
  \begin{subfigure}[b]{0.45\textwidth}
    \includegraphics[width=1.05\linewidth]{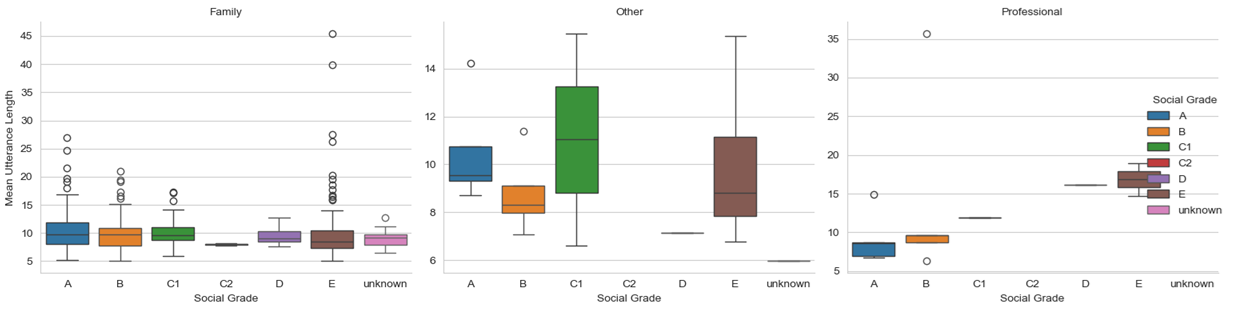}
    \caption{Social Grade and Mean Utterance Length}
    \label{fig:social_grade_utterance_length}
  \end{subfigure}\hfill
  \begin{subfigure}[b]{0.45\textwidth}
    \includegraphics[width=1.05\linewidth]{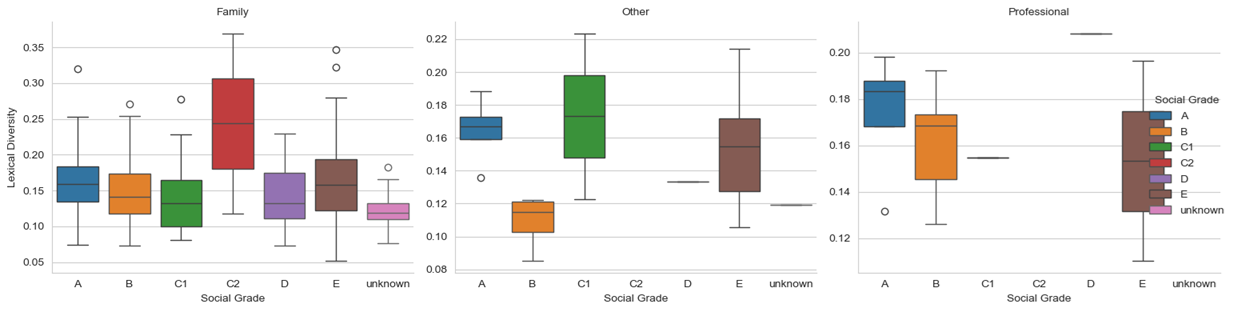}
    \caption{Social Grade and Lexical Diversity}
    \label{fig:social_grade_lexical}
  \end{subfigure}\hfill
  \caption{Dialogue Complexity by (a) Social Grade and Mean Utterance Length (b) Social Grade and Lexical Diversity}
  \label{fig:social_grade_comparison}
\end{figure}

\begin{figure}[ht]
  \centering
  \begin{subfigure}[a]{0.32\textwidth}
    \includegraphics[width=\linewidth]{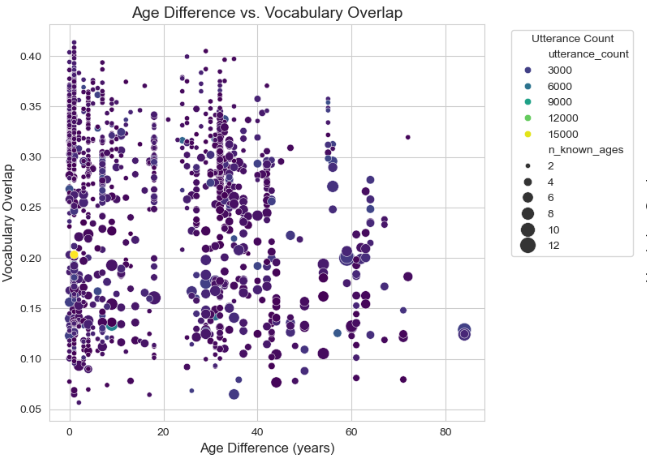}
    \caption{Dialogue Age Diff. vs Vocab Overlap}
    \label{fig:vocab_age_diff}
  \end{subfigure}\hfill
  \begin{subfigure}[b]{0.32\textwidth}
    \includegraphics[width=\linewidth]{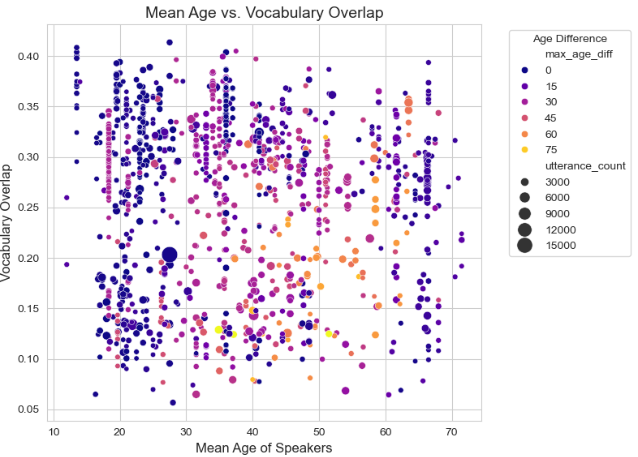}
    \caption{Dialogue Mean Age vs Vocab Overlap}
    \label{fig:vocab_mean_age}
  \end{subfigure}\hfill
  \caption{(a) Dialogue Age Difference vs Vocab Overlap (b) Dialogue Mean Age vs Vocab Overlap}
  \label{fig:vocab_overlap}
\end{figure}

Lexical diversity varied by socio-economic grade in Figure \ref{fig:social_grade_comparison} $(F(6, 482)=5.23, p<.001)$, but not by the context $(F(2, 482)=0.73, p=.39)$ and the interaction was non-significant $(F(12, 482)=0.59, p=0.81)$. In contrast, we observe that vocabulary overlap decreased modestly but significantly as the age difference increased in Figure \ref{fig:vocab_age_diff}. A one-way ANOVA confirms significant differences across age difference categories $(F(4, 968)=8.20, p<.001)$ and correlation shows a modest negative association $(r=-0.219, p<.001)$, whereas in Figure \ref{fig:vocab_mean_age} shows no significant relationship between mean speaker age and lexical alignment $(r= -0.042, p=.16)$. 

%% file: sections/subsections/sub_method.tex
\subsection{Methodologies}

\paragraph{Topic Patterns}
By adopting a two‐stage statistical framework, we attempt to uncover lexemes that distinguish speaker age cohorts in conversational data. First, we apply $\chi^2$ (chi-squared) tests of independence to evaluate the association between individual vocabulary items and predefined age groups, isolating terms that are significantly overrepresented within each demographic segment. Second, we replicate this analysis within each age cohort, stratifying by speaker gender to examine how gender identity modulates term usage, thereby revealing nuanced patterns of socio-demographic variation in language use.

\paragraph{Clustering}
We extend our analysis to semantic patterns by encoding utterances with BERT-based embeddings \citep{devlin2019bert} to capture contextual meaning, then applying UMAP \citep{mcinnes2020umap} dimensionality reduction to visualize the semantic space in two dimensions. Finally, we perform K-means clustering \citep{MacQueen1967} to select the optimal cluster count via silhouette score \citep{ROUS1987} optimization to examine whether language use develops along a continuous spectrum or exhibits distinct transitions between life stages, allowing us to identify insights into naturally occurring language divisions.

\paragraph{Age Classification}
The core classification task employs a comparative approach across multiple models from Scikit-learn \citep{pedregosa2011scikit}. We implemented a baseline OneVsRest Logistic Regression, alongside more complex algorithms including Random Forest with 100 estimators, Gradient Boosting at the learning rate of 0.1 and max depth 5, Linear SVM, and Multi-Layer Perceptron with three hidden layers (128-64-32 neurons). To resolve the class imbalance in age group distribution, we employed Synthetic Minority Oversampling Technique (SMOTE) \citep{Chawla_2002smote} which boosts representation of undersampled age groups while limiting any bias towards majority classes. The feature engineering framework includes four major categories:  (1) textual features extracted through TF-IDF vectorization; (2) lexical features, including type-token ratio and average word length; (3) syntactic features, encompassing utterance length statistics, POS tag distributions, and function word ratios; and (4) discourse and pragmatic features, characterizing hesitation markers, discourse markers, and formality scores.

\subsection{Experimental Approach}

The study employed a stratified demographic categorization framework with participants classified into discrete age cohorts as described in section \ref{subsec:data}. Data pretreatment involves the systematic elimination of uncategorizable occurrences lacking age metadata to assure demographic validity. Text representations employed TF-IDF vectorization with dimensionality confined to 15000 features, imposing a minimum document frequency criterion of 3 while eliminating stop words.

For clustering analysis, we initially balanced the corpus using stratified random sampling of 500 utterances per age cohort. 1688 utterances are left after employing quality control criteria maintaining solely non-null textual occurrences consisting minimally three lexical units. Semantic encoding standardized all utterances to consistent 128-token sequences by truncation and padding techniques, allowing dimensional consistency for later distributional analysis.

Aside from that, all classification experiments were carried out using the combined utterances of every speaker at the speaker level. To ensure methodological consistency and repeatability of results, a fixed random seed of 42 is applied consistently across all experimental components, including classifier initialisation, SMOTE-based oversampling, dataset partitioning, and clustering algorithms.

%% file: sections/results.tex
\section{Results}

\input{sections/subsections/sub_result_topic}
\input{sections/subsections/sub_result_cluster}
\input{sections/subsections/sub_result_classification}

%% file: sections/subsections/sub_result_topic.tex
\subsection{Topic Patterns}

\begin{figure}[ht]
  \centering
  \includegraphics[width=0.6\linewidth]{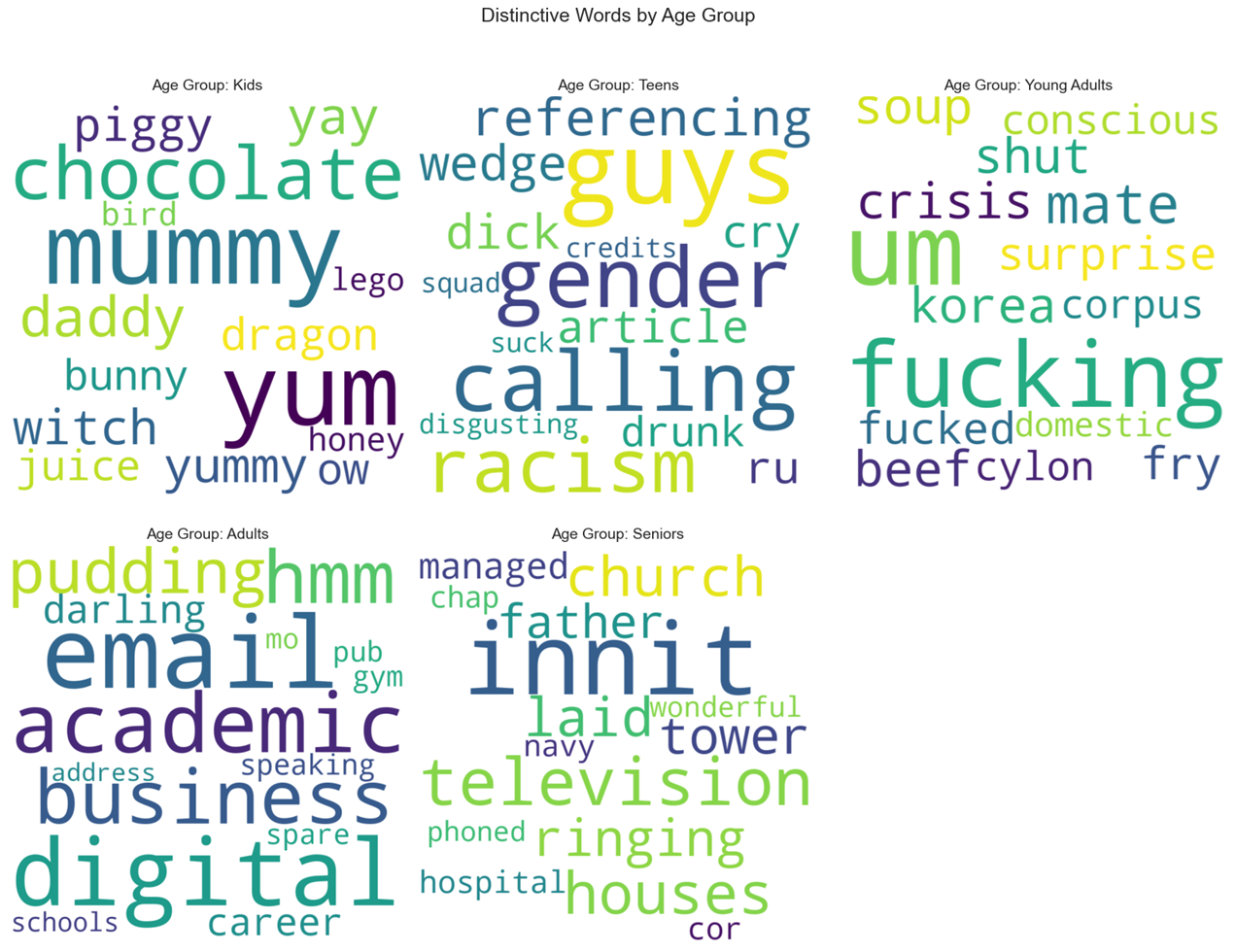}
  \caption{Distinctive Words by Age Group}
  \label{fig:age_bin_words}
\end{figure}

We notice a clear disparity in topical language usage across various age groups as seen from Figure \ref{fig:age_bin_words}. Children’s speech is generally concentrated around themes of family and food, evident in frequent usage of tokens such as mummy, chocolate, and tasty, which imply a concentration on immediate familial and sensory experiences. In contrast, the linguistic patterns observed among teenagers and young adults reveal a greater engagement of more provocative topic matter with often use of explicit language, with profanity and themes such as racism. This signifies a time of heightened emotional expressiveness and social identity construction. Adult speakers exhibit a more utilitarian and occupationally driven lexicon, frequently employing terms such as email, business, and digital, which are indicative of professional discourse practices. Finally, the discourse of seniors reflect a shift toward topics associated with health, religion, and passive media consumption, with recurrent use of tokens such as church, television, and hospital.

%% file: sections/subsections/sub_result_cluster.tex
\subsection{Clustering}

\begin{table}[h!]
\centering
\resizebox{\columnwidth}{!}{%
\begin{tabular}{lcc}
\toprule
\textbf{Feature} & \textbf{Cluster 0 (539 utterances)} & \textbf{Cluster 1 (1149 utterances)} \\
\midrule
\textbf{Age Group Distribution} & & \\
\quad Kids & 13.2\% & 21.8\% \\
\quad Teens & 23.6\% & 21.4\% \\
\quad Young Adults & 22.6\% & 19.6\% \\
\quad Adults & 21.7\% & 19.2\% \\
\quad Seniors & 18.9\% & 17.9\% \\
\addlinespace
\textbf{Most Common Tokens} &
\begin{tabular}[t]{@{}l@{}}%
\textit{like} (240), \textit{yeah} (171), \textit{know} (95),\\
\textit{think} (83), \textit{got} (75), \textit{erm} (74)
\end{tabular} &
\begin{tabular}[t]{@{}l@{}}%
\textit{yeah} (119), \textit{like} (94), \textit{oh} (75),\\
\textit{know} (59), \textit{got} (42), \textit{one} (40)
\end{tabular} \\
\bottomrule
\end{tabular}%
}
\caption{Comparison of Age Group Distribution and Most Common Tokens Across Clusters (see Figure \ref{fig:k_means_cluster})}
\label{tab:cluster_comparison}
\end{table}

The frequent occurrence of discourse markers and hesitation forms such as like, yeah, know, erm, and well in both clusters points to the pervasive use of features associated with spoken, informal interaction. While there are minor variations in lexical frequency across clusters, such as slightly higher use of evaluative terms like backchanneling tokens like mm in one cluster, these differences do not appear to correspond neatly to age. However, in Cosine Similarity matrix (Figure \ref{fig:cosine_age_groups}) we discover similar age groups (e.g., Teens and Young Adults) demonstrated higher similarity scores compared to non-adjacent groups (e.g., Kids and Seniors), suggesting a progressive language shift across the lifespan rather than dramatic alterations.

%% file: sections/subsections/sub_result_classification.tex
\subsection{Age Classification}

Table \ref{tab:perf-by-feature} reveals a clear hierarchy of feature‐group utility and model efficacy. Lexical features alone performed near chance with accuracy ranges between 0.30–0.39 across all models, and the addition of syntactic structure yield modest improvements, peaking at 0.50 with Random Forest. Incorporating discourse features yields a substantial improvement (0.48–0.62), but the strongest standalone predictors arises from raw text embeddings with Linear SVM achieving $\approx$0.73 accuracy and an F1 of $\approx$0.71, substantially outstripping all other model and feature combinations. Notably, concatenating all feature classes fails to improve upon the text-only baseline with the highest accuracy at $\approx$0.71 with Linear SVM, indicating that handcrafted lexical, syntactic, and discourse features contain minimal unique signal beyond that already captured by dense text representations. Across classifiers, Linear SVMs outperform MLP, Random Forest, Gradient Boosting, and Logistic Regression most of the time, suggesting that the demographic distinctions in conversational speech data are largely linearly separable in embedding space.

\begin{table}[ht]
  \centering
  \resizebox{\columnwidth}{!}{%
    \begin{tabular}{llcccc}
      \toprule
      \textbf{Feature Group} & \textbf{Model}            & \textbf{Accuracy} & \textbf{F1 Score} & \textbf{Precision} & \textbf{Recall} \\
      \midrule

      \multirow{5}{*}{Lexical Features}
        & Logistic Regression     & \textbf{0.3962} & 0.2249 & 0.1570 & \textbf{0.3962} \\
        & Random Forest           & 0.3019 & \textbf{0.2908} & \textbf{0.2874} & 0.3019 \\
        & Gradient Boosting       & 0.3019 & 0.2881 & 0.2830 & 0.3019 \\
        & Linear SVM              & \textbf{0.3962} & 0.2249 & 0.1570 & \textbf{0.3962} \\
        & MLP          & \textbf{0.3962} & 0.2644 & 0.2411 & \textbf{0.3962} \\
      \midrule

      \multirow{5}{*}{Syntactic Features}
        & Logistic Regression     & 0.4340 & 0.3715 & 0.3768 & 0.4340 \\
        & Random Forest           & \textbf{0.5000} & \textbf{0.4559} & \textbf{0.4321} & \textbf{0.5000} \\
        & Gradient Boosting       & 0.4340 & 0.3982 & 0.3794 & 0.4340 \\
        & Linear SVM              & 0.4340 & 0.3634 & 0.3715 & 0.4340 \\
        & MLP          & 0.4623 & 0.4057 & 0.3888 & 0.4623 \\
      \midrule

      \multirow{5}{*}{Discourse Features}
        & Logistic Regression     & 0.5943 & 0.5325 & 0.5284 & 0.5943 \\
        & Random Forest           & 0.5660 & 0.5383 & 0.5263 & 0.5660 \\
        & Gradient Boosting       & 0.4811 & 0.4699 & 0.4635 & 0.4811 \\
        & Linear SVM              & 0.5943 & 0.5263 & 0.5340 & 0.5943 \\
        & MLP          & \textbf{0.6226} & \textbf{0.5760} & \textbf{0.5432} & \textbf{0.6226} \\
      \midrule

      \multirow{5}{*}{Text Features}
        & Logistic Regression     & 0.6415 & 0.5860 & 0.5708 & 0.6415 \\
        & Random Forest           & 0.6226 & 0.5667 & 0.5589 & 0.6226 \\
        & Gradient Boosting       & 0.6226 & 0.5771 & 0.5470 & 0.6226 \\
        & Linear SVM              & \textbf{0.7264} & \textbf{0.7080} & \textbf{0.7130} & \textbf{0.7264} \\
        & MLP          & 0.6981 & 0.6496 & 0.6361 & 0.6981 \\
      \midrule

      \multirow{5}{*}{All Features}
        & Logistic Regression     & 0.6415 & 0.5946 & 0.5657 & 0.6415 \\
        & Random Forest           & 0.6321 & 0.5769 & 0.5607 & 0.6321 \\
        & Gradient Boosting       & 0.6226 & 0.5773 & 0.5503 & 0.6226 \\
        & Linear SVM              & \textbf{0.7075} & \textbf{0.6969} & \textbf{0.7088} & \textbf{0.7075} \\
        & MLP          & 0.5849 & 0.5345 & 0.5469 & 0.5849 \\
      \bottomrule
    \end{tabular}%
  }
  \caption{Model Performance by Feature Group (Figure \ref{fig:model_perf_feature})}
  \label{tab:perf-by-feature}
\end{table}

%% file: sections/discussion.tex
\section{Discussion \& Conclusion}



\paragraph{Classification Performance}

The comparative evaluation of classification models yielded several insights regarding the relationship between linguistic features and age prediction. Linear SVM consistently outperformed other classifiers across feature sets $\approx$0.73 accuracy with text features, likely due to its capacity to establish optimal decision boundaries in high-dimensional space. This suggests that age-differentiating linguistic patterns may be linearly separable in feature space.
Among the feature categories, text features proved most predictive, substantially outperforming isolated discourse features, syntactic features, and lexical features. The limited additional benefit from combining all feature types suggests potential redundancy across feature sets, although feature selection technique like K-Best is applied. This finding indicates that basic text features capture most age-related variation, challenging the common assumptions about the necessity of complex feature engineering for sociolinguistic classification.

\paragraph{Lexical Cues from Classifier}
Although the logistic classifier yielded lower accuracy, the resulting feature weights in Figure \ref{fig:top_feature} nonetheless reflect lexical distributions across cohorts that are broadly comparable to those identified by the $\chi^2$ analysis. Caregiver and treat terms like mummy, chocolate, and ice-cream signals a domestic–sensorial focus among kids. Teens load most heavily on peer-oriented stance markers and mild taboos (yeah like, stuff, religion), reflecting evaluative negotiation of identity. Young adults are characterised by hedges and strong intensifiers (mate, fucking, fuck), whereas Adults are signaled by pragmatic planners linked to routine and obligation (work, need, gonna, look). Seniors, in contrast, favour turn-taking signals and epistemic softeners (right yeah, oh right, lovely), aligning with conversational caution observed in later life stages.

\paragraph{Feature Space Continuity of Age-Related Variation}

\begin{figure}[ht]
  \centering
  \includegraphics[width=\columnwidth]{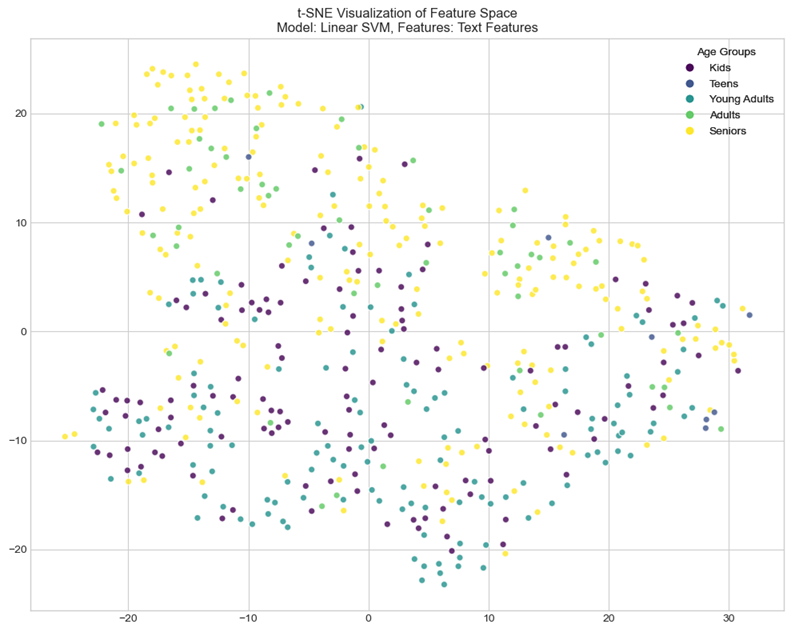}
  \caption{t-SNE Visualization of Feature Space with Linear SVM}
  \label{fig:t_SNE}
\end{figure}

\begin{figure}[ht]
  \centering
  \includegraphics[width=\columnwidth]{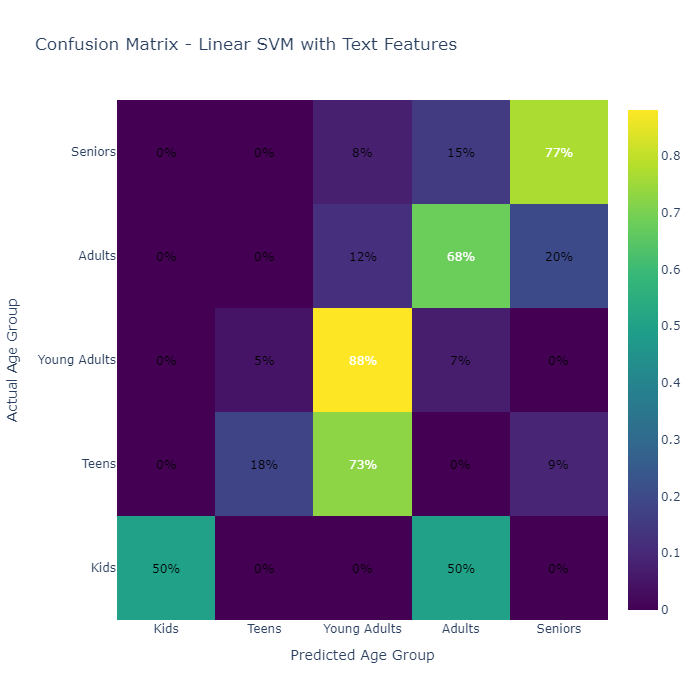}
  \caption{Confusion Matrix for Linear SVM Model with Text Features}
  \label{fig:confusion_matrix}
\end{figure}

The t-SNE projection \citep{vanDerMaaten2008} of the text-feature embeddings in Figure \ref{fig:t_SNE} illustrates age as a continuous manifold instead of a collection of distinct and well-delimited clusters. Observations drawn from the youngest and oldest cohorts occupy antipodal regions of the map, yet the intervening points associated with teens, young adults, and adults form an unbroken chromatic gradient that bridges those extremes. This graded topology implies that the linguistic correlates of chronological age evolve incrementally across the lifespan with no sharp topological discontinuities demarcate neighbouring cohorts. The visual pattern accords with the confusion matrix (Figure \ref{fig:confusion_matrix}), in which misclassifications are concentrated at cohort boundaries, further supporting a life-span continuity model of conversational English and underscoring the empirical porosity of rigid age categories in supervised classification tasks.

\paragraph{Methodological Considerations and Limitations}

Although the BNC2014 dataset represents diverse British speech, its sampling frame exhibits biases that constrain the study’s external validity. Certain demographic strata are over-represented, inflating the lexical and stylistic signals of those groups while attenuating others. Moreover, the substantial variation in words per speaker as seen in Figure \ref{fig:corpus_dist} may also influence speaker-level analyses. Despite SMOTE balancing was employed during training, classification performance varied considerably across age groups, with the highest accuracy for young adults but frequent misclassifications with other categories as observed in Figure \ref{fig:confusion_matrix}. This outcome reflects the underlying age distribution of the corpus, whose median speaker age (34 years) falls squarely within the young-adult category and thus reinforces model bias toward that group.

\subsection{Conclusion}
Across descriptive and predictive lenses, age emerges as the strongest organiser of conversational style: utterances lengthen, vocabularies widen and topical foci shift gradually from childhood to old age. A Linear SVM trained on TF-IDF features recovers five broad age cohorts with 0.73 accuracy, and t-SNE projections reveal a smooth gradient rather than discrete generational islands, confirming that age-related variation is largely continuous in feature space. Socio-economic status and gender modulate, but do not eclipse, this life-span trajectory. Corpus skew towards young, highly educated speakers and the absence of acoustic cues temper the generalisability of these findings, yet the results demonstrate that even basic text representations encode robust demographic signal.

\bigskip
\noindent Word Count: 1990 words

%% file: sections/appendix.tex
\appendix

\section{Appendix}
\label{sec:appendix}



\begin{figure}[ht]
    \centering
    \includegraphics[width=\columnwidth]{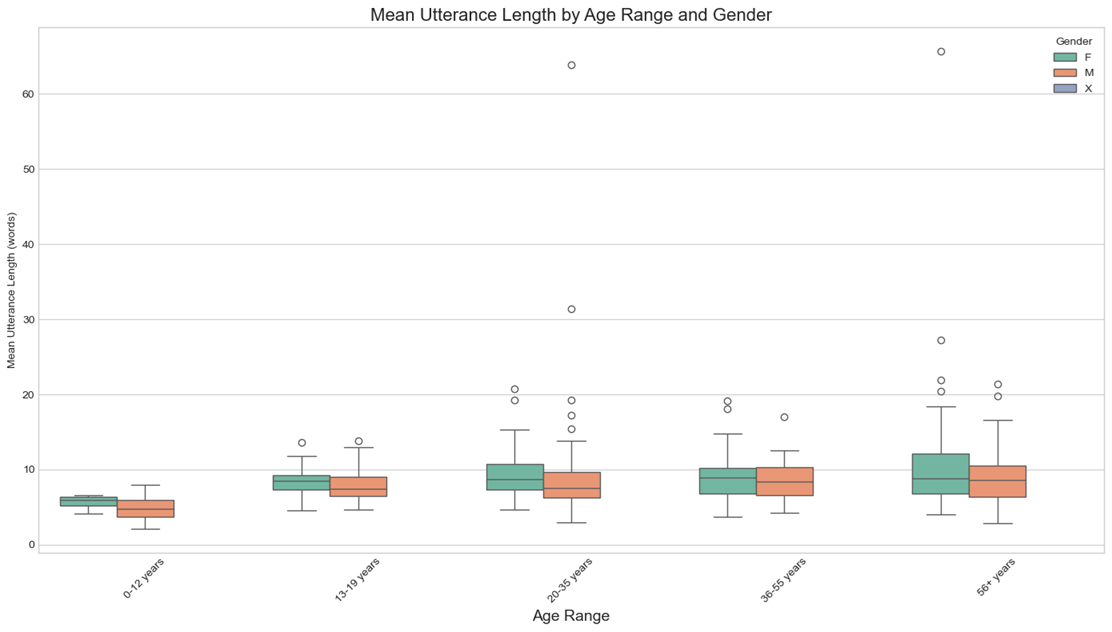}
    \caption{Mean Utterance Length by Age Bins and Gender}
    \label{fig:mean_length_age_gender}
\end{figure}

\begin{figure}[ht]
  \centering
  \includegraphics[width=0.85\columnwidth]{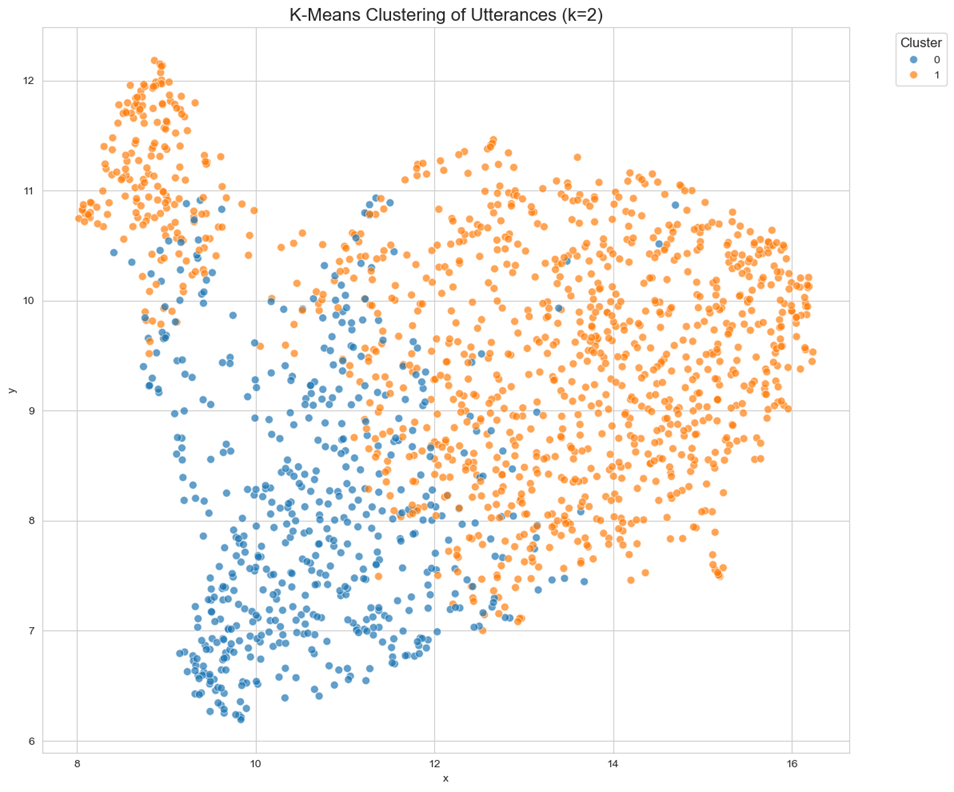}
  \caption{K-Means Clustering of Utterances (k=2)}
  \label{fig:k_means_cluster}
\end{figure}

\begin{figure}[ht]
    \centering
    \includegraphics[width=0.85\columnwidth]{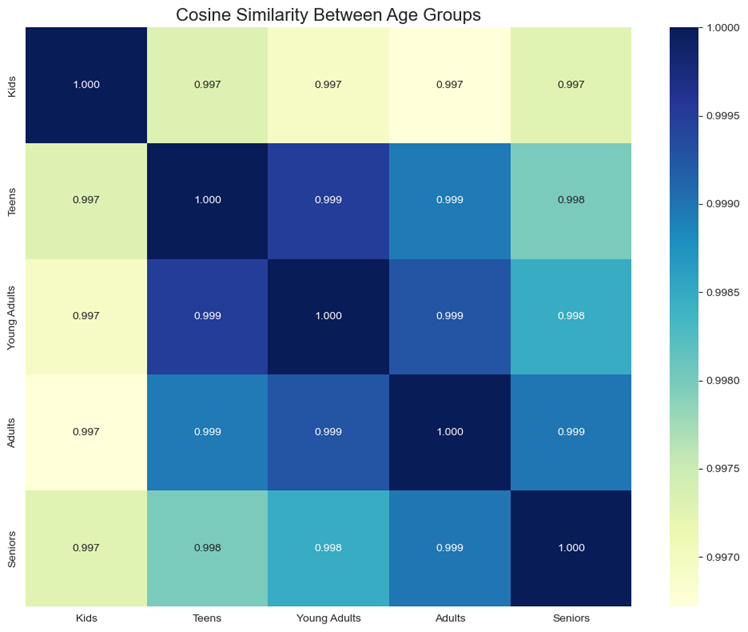}
    \caption{Cosine Similarity Between Age Groups}
    \label{fig:cosine_age_groups}
\end{figure}

\begin{figure}[ht]
  \centering
  \includegraphics[width=0.85\columnwidth]{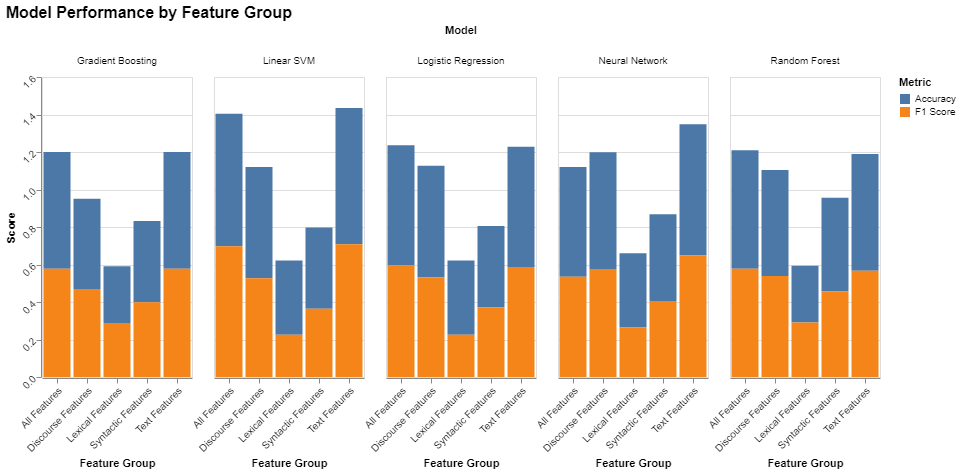}
  \caption{Model Performance Comparison Bar Plot by Feature Group}
  \label{fig:model_perf_feature}
\end{figure}


\begin{figure}[ht]
  \centering
  \includegraphics[width=0.85\columnwidth]{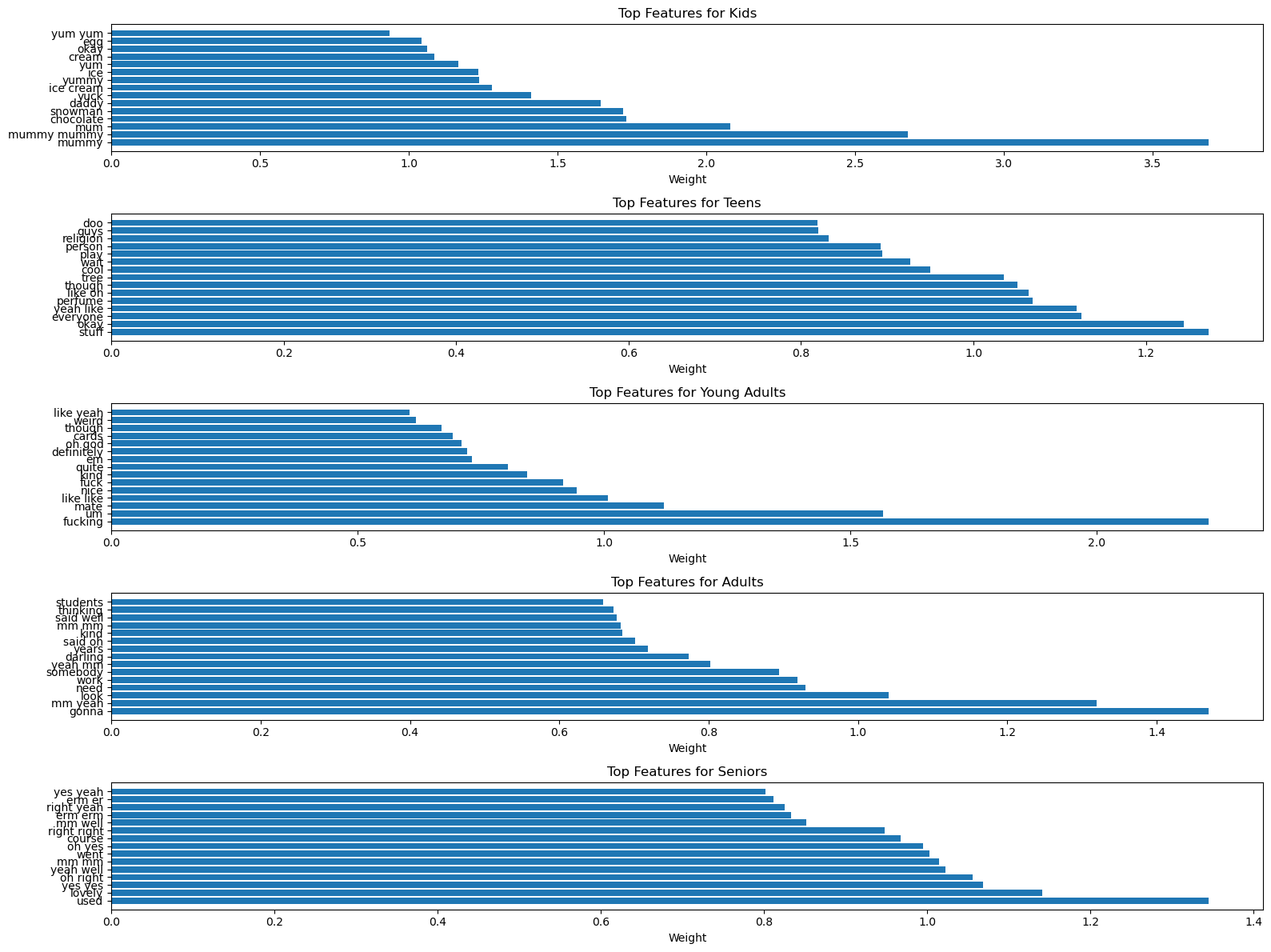}
  \caption{Top Predictive Features for Different Age Groups on One-vs-Rest Logistic Regression Classifier}
  \label{fig:top_feature}
\end{figure}






%% file: main.bbl
\begin{thebibliography}{8}
\expandafter\ifx\csname natexlab\endcsname\relax\def\natexlab#1{#1}\fi

\bibitem[{Chawla et~al.(2002)Chawla, Bowyer, Hall, and Kegelmeyer}]{Chawla_2002smote}
N.~V. Chawla, K.~W. Bowyer, L.~O. Hall, and W.~P. Kegelmeyer. 2002.
\newblock \href {https://doi.org/10.1613/jair.953} {Smote: Synthetic minority over-sampling technique}.
\newblock \emph{Journal of Artificial Intelligence Research}, 16:321–357.

\bibitem[{Devlin et~al.(2019)Devlin, Chang, Lee, and Toutanova}]{devlin2019bert}
Jacob Devlin, Ming-Wei Chang, Kenton Lee, and Kristina Toutanova. 2019.
\newblock \href {http://arxiv.org/abs/1810.04805} {Bert: Pre-training of deep bidirectional transformers for language understanding}.

\bibitem[{Love et~al.(2017)Love, Dembry, Hardie, Brezina, and McEnery}]{BNC2014}
Robbie Love, Claire Dembry, Andrew Hardie, Vaclav Brezina, and Tony McEnery. 2017.
\newblock \href {https://doi.org/10.1075/ijcl.22.3.02lov} {The spoken bnc2014: Designing and building a spoken corpus of everyday conversations}.
\newblock \emph{International Journal of Corpus Linguistics}, 22(3):319--344.
\newblock {\textcopyright} John Benjamins Publishing Company This is an open access article under a OA CC BY license.

\bibitem[{MacQueen(1967)}]{MacQueen1967}
J.~B. MacQueen. 1967.
\newblock Some methods for classification and analysis of multivariate observations.
\newblock In \emph{Proc. of the fifth Berkeley Symposium on Mathematical Statistics and Probability}, volume~1, pages 281--297. University of California Press.

\bibitem[{McInnes et~al.(2020)McInnes, Healy, and Melville}]{mcinnes2020umap}
Leland McInnes, John Healy, and James Melville. 2020.
\newblock \href {http://arxiv.org/abs/1802.03426} {Umap: Uniform manifold approximation and projection for dimension reduction}.

\bibitem[{Pedregosa et~al.(2011)Pedregosa, Varoquaux, Gramfort, Michel, Thirion, Grisel, Blondel, Prettenhofer, Weiss, Dubourg et~al.}]{pedregosa2011scikit}
Fabian Pedregosa, Ga{\"e}l Varoquaux, Alexandre Gramfort, Vincent Michel, Bertrand Thirion, Olivier Grisel, Mathieu Blondel, Peter Prettenhofer, Ron Weiss, Vincent Dubourg, et~al. 2011.
\newblock Scikit-learn: Machine learning in python.
\newblock \emph{Journal of machine learning research}, 12(Oct):2825--2830.

\bibitem[{Rousseeuw(1987)}]{ROUS1987}
P.~Rousseeuw. 1987.
\newblock Silhouette: a graphical aid to the interpretation and validation of cluster analysis.
\newblock \emph{Journal of Computational and Applied Mathematics}, 20:53--65.

\bibitem[{van~der Maaten and Hinton(2008)}]{vanDerMaaten2008}
Laurens van~der Maaten and Geoffrey Hinton. 2008.
\newblock \href {http://www.jmlr.org/papers/v9/vandermaaten08a.html} {Visualizing data using {t-SNE}}.
\newblock \emph{Journal of Machine Learning Research}, 9:2579--2605.

\end{thebibliography}
